\documentclass[11pt,a4paper]{article}
\usepackage[hyperref]{eacl2021}
\usepackage{times}
\usepackage{latexsym}

\usepackage{graphicx}

\usepackage{microtype}
\usepackage{multirow}
\usepackage{bbm}

\usepackage{amsmath,amsfonts,bm}

\def\eqref#1{equation~\ref{#1}}

\def\1{\bm{1}}

\def\vb{{\bm{b}}}

\def\vd{{\bm{d}}}
\def\ve{{\bm{e}}}

\def\vs{{\bm{s}}}

\def\vu{{\bm{u}}}

\def\vw{{\bm{w}}}

\DeclareMathAlphabet{\mathsfit}{\encodingdefault}{\sfdefault}{m}{sl}
\SetMathAlphabet{\mathsfit}{bold}{\encodingdefault}{\sfdefault}{bx}{n}

\DeclareMathOperator*{\argmin}{arg\,min}

\def\lessgreat#1{\textless {#1}\textgreater{}}

\usepackage[ruled,vlined]{algorithm2e}

\aclfinalcopy 
\def\aclpaperid{355} 

\newcommand{\refAppendixAuto}{the appendix}
\newcommand{\refAppendixMethod}{the appendix}
\newcommand{\refAppendixEntail}{the appendix}

\title{Multi-facet Universal Schema}

\author{Rohan Paul\thanks{* indicates equal contribution} \ \ Haw-Shiuan Chang\footnotemark[1] \ \ Andrew McCallum \\
  CICS, University of Massachusetts Amherst \\
  \texttt{\{rohpaul,hschang,mccallum\}@cs.umass.edu} \\
}

\date{}

\begin{document}
\maketitle
\begin{abstract}
Universal schema (USchema) assumes that two sentence patterns that share the same entity pairs are similar to each other. This assumption is widely adopted for solving various types of relation extraction (RE) tasks. Nevertheless, each sentence pattern could contain multiple facets, and not every facet is similar to all the facets of another sentence pattern co-occurring with the same entity pair. To address the violation of the USchema assumption, we propose multi-facet universal schema that uses a neural model to represent each sentence pattern as multiple facet embeddings and encourage one of these facet embeddings to be close to that of another sentence pattern if they co-occur with the same entity pair. In our experiments, we demonstrate that multi-facet embeddings significantly outperform their single-facet embedding counterpart, compositional universal schema (CUSchema)~\citep{verga2016multilingual}, in distantly supervised relation extraction tasks. Moreover, we can also use multiple embeddings to detect the entailment relation between two sentence patterns when no manual label is available. 



\end{abstract}

\section{Introduction}

Relation extraction (RE) is a crucial step in automatic knowledge base construction (AKBC). A major challenge of RE is that the frequency of relations in the real world is a long-tail distribution but collecting sufficient human annotations for every relation is infeasible~\citep{han2020more}. 


Distant supervision is proposed to alleviate the issue~\citep{mintz2009distant}. Distant supervision assumes that a sentence pattern expresses a relation if the sentence pattern
co-occurs with an entity pair and the entity pair has the relation. For example, we assume the sentence pattern \textit{``\$ARG1, the partner of fellow \$ARG2''} is likely to express the spouse relation if we observe a text clip \textit{``... Angelina Jolie, the partner of fellow Brad Pitt ...''} in our training corpus and a knowledge base tells us that \textit{Angelina Jolie} and \textit{Brad Pitt} has the spouse relation.
Accordingly, we can infer that another entity pair is likely to have the spouse relation if we observe the text \textit{``, the partner of fellow''} between them in a new corpus.

\begin{figure}[t!]
\centering
\includegraphics[width=1\linewidth]{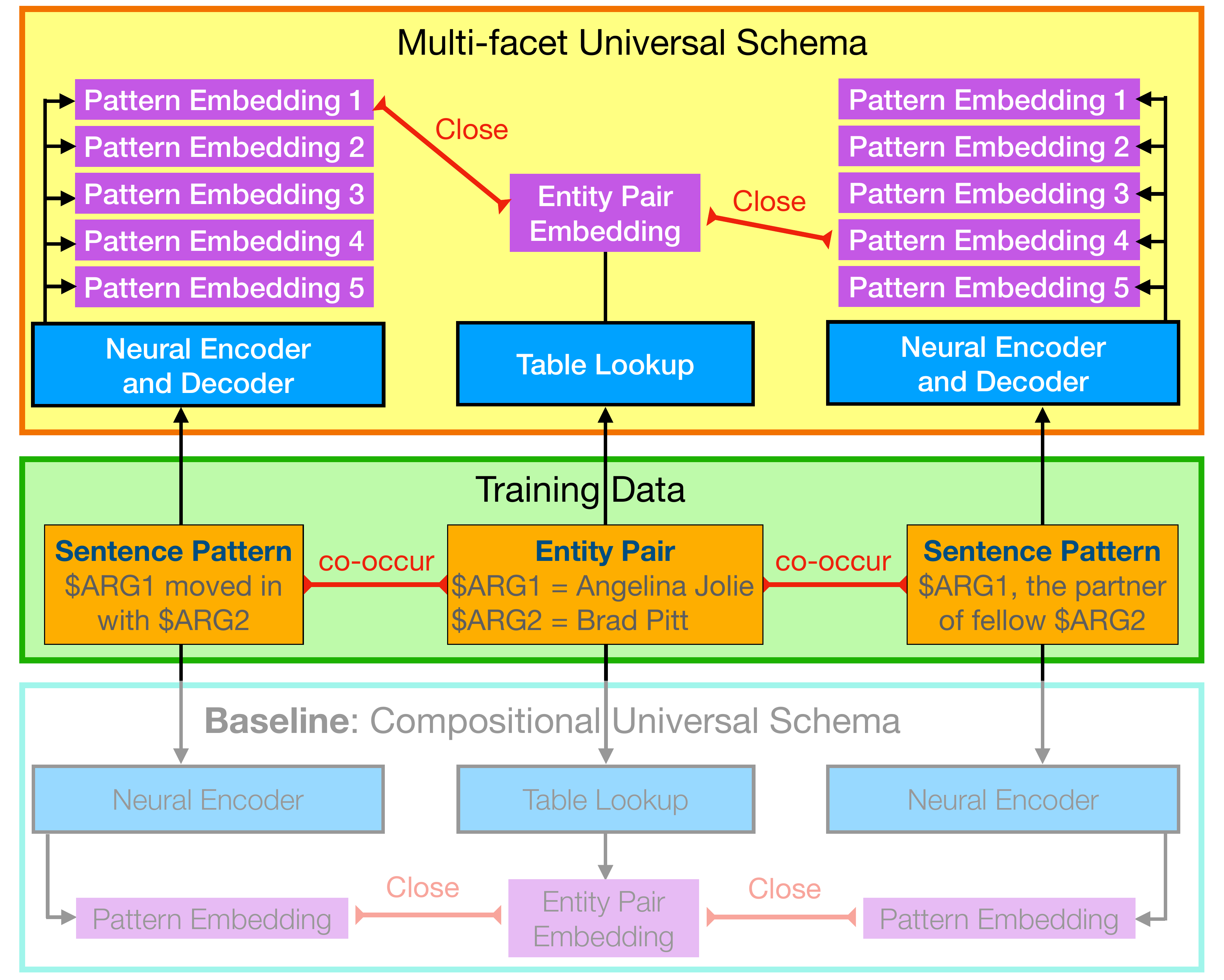}
\caption{Comparison between the multi-facet and compositional universal schema. In our training loss, we encourage one of the facet embeddings from a sentence pattern to be similar to its co-occurred entity pair.}
\label{fig:first_page}
\end{figure}


Universal schema~\citep{riedel2013relation} extends this assumption by treating every sentence pattern as a relation, which means we assume that sentence patterns or relations in a knowledge base are similar if they co-occur with the same entity pair. For example, we assume \textit{``\$ARG1, the partner of fellow \$ARG2''} and \textit{``\$ARG1, the wife of fellow \$ARG2''} are similar if they both co-occur with \textit{(Kristen Bell, Dax Shepard)}. Consequently, we can infer that \textit{``\$ARG1, the wife of fellow \$ARG2''} also implies spouse relation as \textit{``\$ARG1, the partner of fellow \$ARG2''} even if the knowledge base does not record the spouse relation between \textit{Kristen Bell} and \textit{Dax Shepard}.

Compositional universal schema~\citep{verga2016multilingual} realizes the idea by using a LSTM~\citep{hochreiter1997long} to encode each sentence pattern into an embedding and encouraging the embedding to be similar to the embedding of the co-occurred entity pair. 
As in the lower part of Figure~\ref{fig:first_page}, the model makes the embeddings of two sentence patterns similar if they co-occur with the same entity pair. \citet{soares2019matching} rely on a similar assumption and achieve state-of-the-art results on supervised RE tasks by replacing the LSTM with a large pre-trained language model.


\begin{figure*}[t!]
\centering
\includegraphics[width=1\linewidth]{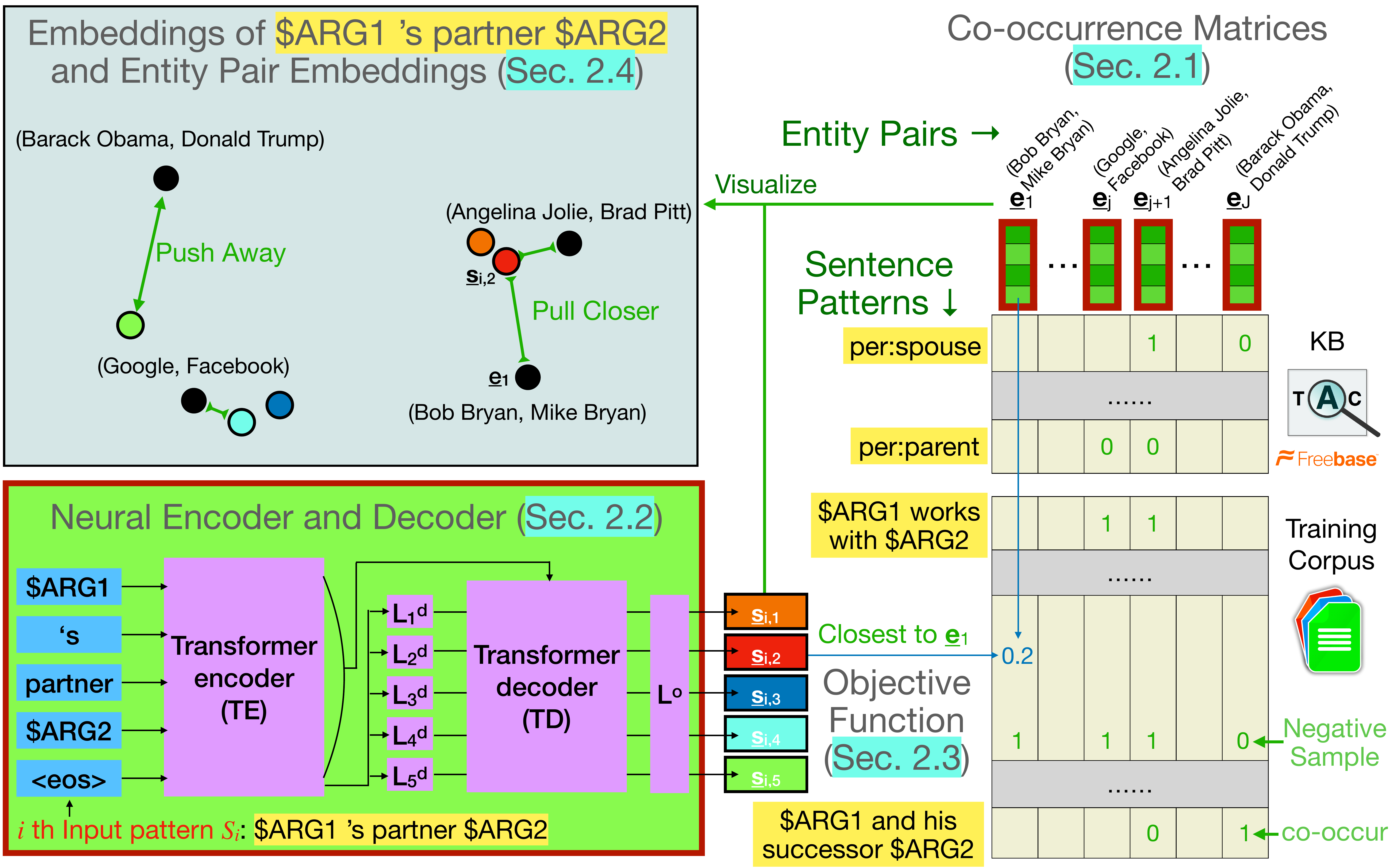}
\caption{An illustration of the proposed method. The training signal comes from the co-occurrence matrices of the KB and training text corpus on the right. On the lower left, we visualize our neural encoder, which captures the compositional meaning of tokens in the sentence pattern, and our neural decoder, which models the dependency among multiple facet embeddings. When a sentence pattern co-occurs with an entity pair, the training loss minimizes the distance between the entity pair embedding and the closest facet embedding of the sentence pattern (e.g., 0.2 between $s_{i,2}$ and $e_{1}$). Trainable parameters in our model are highlighted using red borders.
On the upper left, we visualize the embedding space to establish the connection between our method and clustering. 
}
\label{fig:architecture}
\end{figure*}

The variants of universal schema have many different applications, including multilingual RE~\cite{verga2016multilingual}, knowledge base construction~\cite{toutanova2015representing,verga2017generalizing}, question answering~\cite{das2017question}, document-level RE~\cite{verga2018simultaneously}, N-ary RE~\cite{akimoto2019cross}, open information extraction~\cite{zhang2019openki}, and unsupervised relation discovery~\cite{percha2018global}. 

Nevertheless, one sentence pattern could contain multiple facets, and each facet could imply a different relation. In Figure~\ref{fig:first_page}, \textit{``\$ARG1, the partner of fellow \$ARG2''} could imply the entity pair has the spouse relation, the co-worker relation, or both. \textit{``\$ARG1 moved in with \$ARG2''} could imply the spouse relation, the parent relation, ..., etc. If we squeeze the facets of a sentence pattern into a single embedding, the embedding is more likely to be affected by the irrelevant facets from other patterns co-occurred with the same entity pair (e.g., \textit{``\$ARG1 moved in with \$ARG2''} might incorrectly imply the co-worker relation).

Another limitation is that single embedding representation can only provide symmetric similarity measurement between two sentence patterns. Thus, an open research challenge is to predict the entailment direction of two sentence patterns only based on their co-occurring entity pair information. 

To overcome the challenges, we propose multi-facet universal schema, where we assume that two sentence patterns share a similar facet if they co-occur with the same entity pair. As in Figure~\ref{fig:first_page}, we use a neural encoder and decoder to predict multiple facet embeddings of each sentence pattern and encourage one of the facet embeddings to be similar to the entity pair embedding. As a result, the facets that are irrelevant to the relation between the entity pairs are less likely to affect the embeddings of entity pairs and other related sentence patterns. For example, the parent facet of \textit{``\$ARG1 moved in with \$ARG2''} could be excluded when updating the embeddings of \textit{(Angelina Jolie, Brad Pitt)}.

In our experiments, we first compare the multi-facet embeddings with the single-facet embedding in distantly supervised RE tasks. The results demonstrate that multiple facet embeddings significantly improve the similarity measurement between the sentence patterns and knowledge base relations. Besides RE, we also apply multi-facet embeddings to unsupervised entailment detection tasks. In a newly collected dataset, we show that multi-facet universal schema significantly outperforms the other unsupervised baselines.

\section{Methods}

Our method is illustrated in Figure~\ref{fig:architecture}. In Section~\ref{sec:background}, we first provide our problem setup: We are given a knowledge base (KB) and a text corpus during training. Our goal is to extract relations by measuring the similarity between KB relations and an (unseen) sentence pattern or to detect entailment between two sentence patterns. In Section~\ref{sec:neural}, we introduce our neural model, which predicts multi-facet embeddings of each sentence pattern. Next, in Section~\ref{sec:obj}, we describe our objective function, which encourages the embeddings of co-occurred entity pairs to be close to the embeddings of their closest pattern facets. Finally, in Section~\ref{sec:cluster}, we explain that multi-facet embeddings could be viewed as the cluster centers of possibly co-occurred entity pairs, and in Section~\ref{sec:scoring}, we provide our scoring functions for distantly supervised RE and unsupervised entailment tasks.


\subsection{Background and Problem Setup}
\label{sec:background}
Our RE problem setup is the same as compositional universal schema~\citep{verga2016multilingual}. First, we run named entity recognition (NER) and entity linking on a raw corpus. After identifying the entity pairs in each sentence, we prepare a co-occurrence matrix as in Figure~\ref{fig:architecture}. Similarly, we represent the KB relations between entity pairs as a co-occurrence matrix and merge the matrices from the KB and the training corpus. The merged matrix has $y_{i,j} = 1$ if the $i$th sentence pair or KB relation co-occurs with the $j$th entity pair and $y_{i,j} = 0$ otherwise. 


During testing, we use NER to extract an entity pair and the sentence pattern, which might not have been seen in the training corpus. Next, we extract relations by computing the similarity between the sentence pattern embeddings and the embeddings of the applicable KB relations. Besides RE, we also detect the entailment between two sentence patterns by comparing their embeddings.




\subsection{Neural Encoder and Decoder}
\label{sec:neural}

We use a neural model to predict $K$ facet embeddings of each sentence pattern. The goal is similar to \citet{NSD}, which predict a fixed number of embeddings of a sentence, so we adopt their neural model as shown in Figure~\ref{fig:architecture}.


For the $i$th sentence pattern $S_{i}$, we append an \lessgreat{eos} to its end and use a 3-layer Transformer~\cite{vaswani2017attention} encoder $TE$ to model the compositional meaning of the input word sequence: 
$\underline{\vu}_{i,1} ... \underline{\vu}_{i,|S_i|} \underline{\vu}_{i,\text{\lessgreat{eos}}} = 
TE( S_{i} \text{\lessgreat{eos}} ), $
where $\underline{\vu}_{i,l}$ is an embedding contextualized by the encoder. In the experiment, we also replace the Transformer with a bidirectional LSTM (bi-LSTM) to show that the improvement of multi-facet embeddings is independent of the encoder choice.

The embedding $\underline{\vu}_{i,\text{\lessgreat{eos}}}$ represents the whole sentence pattern; we use $K$ different linear layers $L^d_k$ to transform the embedding into the inputs of our decoder:
$\underline{\vb}_{i,k} = L^d_k(\underline{\vu}_{i,\text{\lessgreat{eos}}})$.


The facets in a sentence pattern often have some dependency. For example, the patterns that express the partnership between two people might also express the collaboration relation between two companies. To leverage the dependency, we use another 3-layer Transformer as our decoder $TD$. Besides the self-attention, we allow the hidden states in the decoder to query the contextualized word embeddings $\underline{\vu}_{i,l}$ from the encoder~\cite{vaswani2017attention} and output the embeddings corresponding to the different facets $\underline{\vd}_{i,k}$: $\underline{\vd}_{i,1} ... \underline{\vd}_{i,K} = TD(\underline{\vb}_{i,1} ..., \underline{\vb}_{i,K}, \underline{\vu}_{i,1} ... \underline{\vu}_{i,\text{\lessgreat{eos}}}).$ Notice that we do not use autoregressive decoding as in~\citet{vaswani2017attention}, so our decoder could also be viewed as another encoder with attention to the output of the encoder $TE$. Finally, to convert the hidden state size to the entity embedding size, we let the outputs of decoder go through another linear layer $L^o$ to get the facet embedding (i.e., sentence pattern embedding): $\underline{\vs}_{i,k} = L^o(\underline{\vd}_{i,k})$.





\subsection{Objective Function}
\label{sec:obj}

When measuring the distance between the $j$th entity pair and the $i$th sentence pattern, we compute the Euclidean distance between the entity pair embedding $\tilde{\underline{\ve}}_j$ and its closest facet embedding of the $i$th sentence pattern. The distance is defined as

\vspace{-3mm}
\small
\begin{align}
& D(\{\underline{\vs}_{i,k}\}_{k=1}^K, \tilde{\underline{\ve}}_j)  = \min_{k=1}^K \min_{0 \leq \eta_k \leq 1}|| \tilde{\underline{\ve}}_j - \eta_k \underline{\vs}_{i,k} ||^2,
\label{eq:dist_max}
\end{align}
\normalsize
\noindent where the entity pair embedding is normalized  (i.e., $||\tilde{\underline{\ve}}_j||=1$). During testing, we ignore the magnitude of facet embeddings, so we use $\eta_k$ to eliminate the magnitude of facet embeddings $\underline{\vs}_{i,k}$ during training. We do not allow negative $\eta_k$ to prevent the gradient flow from pushing $\underline{\vs}_{i,k}$ toward the inverse direction of $\tilde{\underline{\ve}}_j$ and we ensure $\eta_k \leq 1$ to avoid the neural model from outputting $\underline{\vs}_{i,k}$ with a very small magnitude.


As in Figure~\ref{fig:architecture}, we minimize the distance $D(\{\underline{\vs}_{i,k}\}_{k=1}^K, \tilde{\underline{\ve}}_j)$ in our loss function when the $i$th sentence pair co-occurs with the $j$th entity pair (i.e., $y_{i,j} = 1$). For negative samples (i.e., $y_{i,j} = 0$), we maximize the distance instead. That is, the major term of our loss function is defined as

\vspace{-3mm}
\begin{align}
& \sum_{(i,j) \in R} (2 \cdot y_{i,j}-1) r_{i,j}  D(\{\underline{\vs}_{i,k}\}_{k=1}^K, \tilde{\underline{\ve}}_j), 
\label{eq:dist_obj}
\end{align}
\normalsize
\noindent 
and the other regularization term $\Omega$ in the loss function will be described in \refAppendixAuto. $R$ is a set that includes all positive and negative samples. Positive samples are $(i,j)$ such that $y_{i,j} = 1$ and
the negative samples are constructed by pairing a randomly selected sentence pattern with the $j$th entity pair. 
To balance the influence of popular entity pairs (i.e., entity pairs that co-occur with many sentence patterns) and rare entity pairs on our model,
we set the weight of each pair, $r_{i,j} \propto \frac{1}{\sum_i {y_{i,j}}}$ and $\frac{\sum_{(i,j)\in R} r_{i,j}}{|R|} = 1$.

We generate the embeddings for KB relations in a similar way. We use a single token to represent the relation and append an \lessgreat{eos} (e.g., \emph{per:spouse} \lessgreat{eos}) to form the input of our neural model. 
The KB relations usually co-occur with more entity pairs, so we set the number of facet embeddings for KB relations $K_{rel}$ to be larger than the number of facet embeddings for sentence patterns $K$.


\subsection{Connection to Clustering}
\label{sec:cluster}

If a sentence pattern contains multiple facets that describe different relations between the entity pairs, the pattern often co-occurs with different kinds of entity pairs. For example, \textit{``\$ARG1 ’s partner \$ARG2''} in Figure~\ref{fig:architecture} could express the collaboration relationship between two companies or the partnership between two people, so the sentence patterns could co-occur with two companies such as \textit{(Google, Facebook)} and two people such as \textit{(Bob Bryan, Mike Bryan)}.

Different kinds of entity pairs often have very different embeddings, so we could discover the facets of sentence patterns by clustering the embeddings of entity pairs. Here, a facet refers to a mode of the embedding distribution of the entity pairs that could possibly co-occur with the sentence pattern. A facet could be represented by multiple facet embeddings and each facet embedding corresponds to a cluster center of the entity pair embeddings. Hence, although the number of facet embeddings $K$ is fixed for all the sentence patterns, our model can capture the facets of the sentence patterns well when the number of facets is less than $K$.

In \eqref{eq:dist_max}, we choose the closest facet embedding of the sentence pattern for each co-occurring entity pair embedding and minimize their distance. For example, $\underline{\vs}_{i,2}$ and the embedding of \textit{(Bob Bryan, Mike Bryan)} are pulled closer in Figure~\ref{fig:architecture}.
Minimizing \eqref{eq:dist_max} by passing the gradient through the scaled facet embedding $\eta_k \underline{\vs}_{i,k}$ is the same as minimizing a Kmeans loss, so the loss term induced by positive sample pairs encourage each $\underline{\vs}_{i,k}$ to become the cluster center of its nearby co-occurring entity pair embeddings. The details of our training algorithm could be found in \refAppendixMethod.


The co-occurrence matrices in RE tasks are usually extremely sparse, and most of the sentence patterns only co-occur with a few entity pairs, which makes it difficult to derive multiple high-quality embeddings by clustering the co-occurring entity pair embeddings as in multi-sense word embedding 
methods such as~\citet{NeelakantanSPM14}. The proposed method solves this sparsity challenge by predicting the cluster centers using a neural model. For instance, even if \textit{``\$ARG1 's partner \$ARG2''} does not co-occur with many entity pairs, its embeddings are encouraged to be close to the embeddings of entity pairs that co-occur with other similar patterns (e.g., \textit{``\$ARG1 and her partner \$ARG2''}).






\begin{figure}[t!]
\centering
\includegraphics[width=1\linewidth]{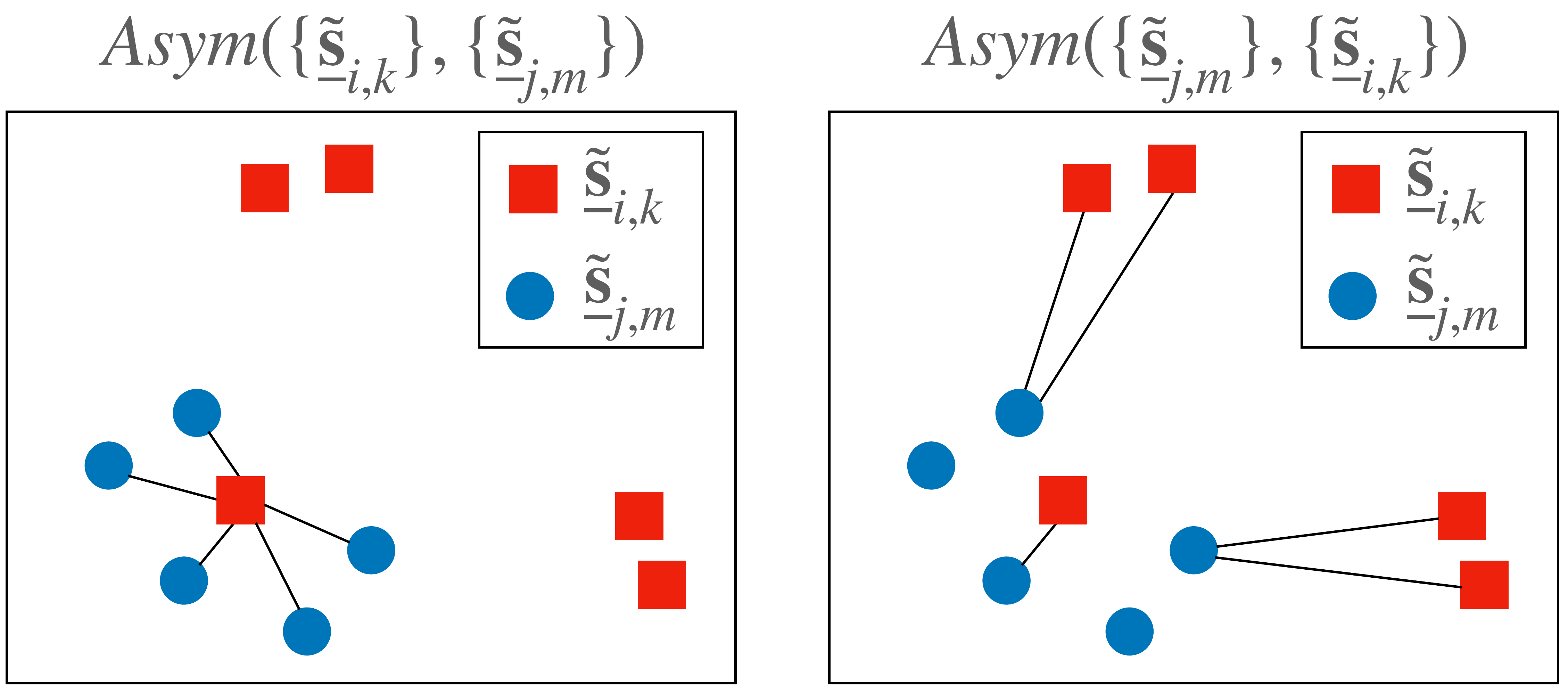}
\caption{Comparison of the asymmetric similarities. $Asym(\{\tilde{\underline{\vs}}_{i,k}\},\{\tilde{\underline{\vs}}_{j,m}\} )>Asym(\{\tilde{\underline{\vs}}_{j,m}\},\{\tilde{\underline{\vs}}_{i,k}\} )$ because the average cosine distance on the left is smaller than that on the right.}
\label{fig:scoring}
\end{figure}

\subsection{Scoring Functions}
\label{sec:scoring}

In compositional universal schema, the similarity between the $i$th and $j$th sentence patterns are measured by the symmetric cosine similarity $\tilde{\underline{\vs}}^T_{i,1}   \tilde{\underline{\vs}}_{j,1}$, where $\tilde{\underline{\vs}}_{i,1} = \frac{\underline{\vs}_{i,1}}{||\underline{\vs}_{i,1}||}$. When using multiple embeddings to represent a sentence pattern, we can compute the asymmetric similarity as 

\vspace{-3mm}
\small
\begin{align}
&  Asym(\{\tilde{\underline{\vs}}_{i,k}\},\{\tilde{\underline{\vs}}_{j,m}\} ) = \frac{\sum_{m=1}^K \max\limits_{k=1}^K(  \tilde{\underline{\vs}}^T_{i,k}   \tilde{\underline{\vs}}_{j,m}  )}{K}.
\end{align}
\label{eq:asym_score}
\normalsize
\noindent In an example of Figure~\ref{fig:scoring}, a red square $\tilde{\underline{\vs}}_{i,k}$ is close to all the blue points, which leads to a high $Asym(\{\tilde{\underline{\vs}}_{i,k}\},\{\tilde{\underline{\vs}}_{j,m}\} )$.

Between two sentence patterns with entailment relation, we empirically find that the embeddings of a premise (the more specific pattern) often have some facet embeddings that are far away from all the embeddings of its hypothesis (the more general pattern). Relying on the tendency, we could detect the direction of the entailment relation. For example, the $i$th sentence pattern (red squares) in Figure~\ref{fig:scoring} is more likely to be premise if the $i$th and $j$th (blue circles) sentence patterns have an entailment relation. 

We suspect the reason is that more specific patterns could contain more words that are similar to the words of other patterns expressing different relations.
For example, \textit{``\$ARG1 , the wife of fellow \$ARG2''} have a facet embedding for spouse relation and another facet embedding for the co-worker relation because the pattern has high word overlapping with \textit{``\$ARG1 , the wife of \$ARG2''} and \textit{``\$ARG1 and her fellow \$ARG2''}. Another possible reason is that the articles in our corpus tend to use more specific patterns to express the relation between a pair of entities~\cite{shwartz2017hypernyms}.






When performing RE, we compute the symmetric similarity between $i$th sentence pattern and $j$th KB relation $Sim(\{\tilde{\underline{\vs}}_{i,k}\},\{\tilde{\underline{\vs}}_{j,m}\} )$ by

\vspace{-3mm}
\small
\begin{align}
&  \frac{ Asym(\{\tilde{\underline{\vs}}_{i,k}\},\{\tilde{\underline{\vs}}_{j,m}\} ) + Asym(\{\tilde{\underline{\vs}}_{j,m}\}, \{\tilde{\underline{\vs}}_{i,k}\} ) }{2}.
\label{eq:sim_score}
\end{align}
\normalsize






\section{Experiments}

We primarily compare our method with compositional universal schema (CUSchema)~\citep{verga2016multilingual} because CUSchema is one of the state-of-the-art RE methods in the small model regime (without using large pre-trained language models)~\cite{chang2016extracting,chaganty2017importance}.\footnote{We have not yet applied the multi-facet embeddings approach to the models that rely on a large pretrained language model (LM)~\cite{soares2019matching} due to computational and evaluation considerations. Computationally speaking, training state-of-the-art models requires intensive GPU resources. Besides, a smaller model size might be desired when we need to construct a knowledge base from a large corpus in real time.
Moreover, there is no existing pretrained LM in some domains~\cite{zhang2019openki}, and training the LM in a new domain from scratch requires even more GPU resources. 

In terms of the evaluation consideration, our method is an improvement over CUSchema, so we want to compare it with CUSchema fairly. Furthermore, evaluating entailment between two full sentences is more difficult than between the sentence patterns, and we are not aware of a LM-based model that only considers the text between the entity pairs.}

In Section~\ref{sec:visual}, we visualize and analyze the facet embeddings. Next, we use distant-supervised RE tasks to evaluate our symmetric similarity measurement in Section~\ref{sec:RE_eval}, and detect entailment between sentence patterns to evaluate our asymmetric similarity measurement in Section~\ref{sec:entail_eval}. 


\begin{figure*}[t!]
\centering
\includegraphics[width=1\linewidth]{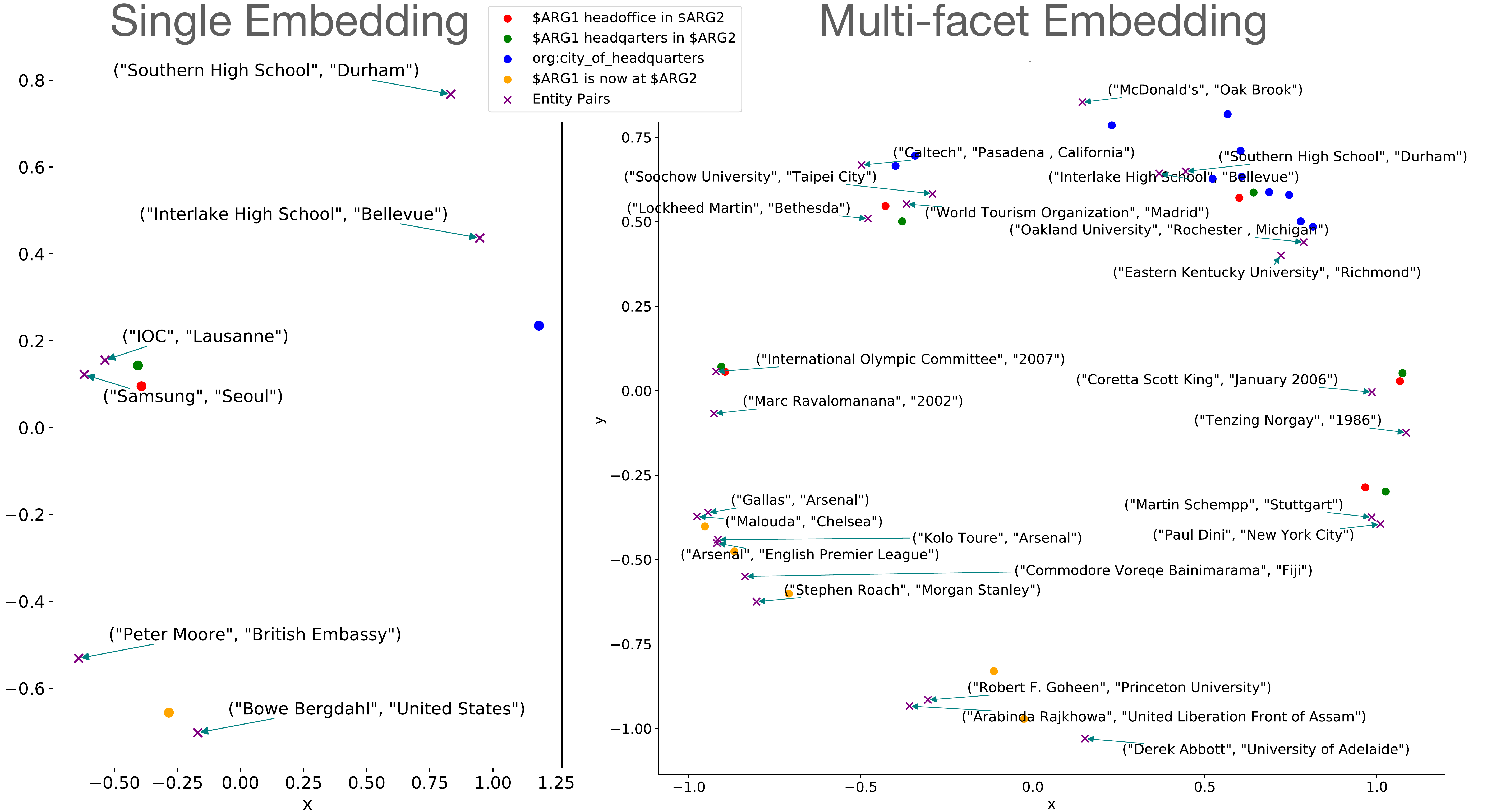}
\caption{Facet embedding visualization of \textbf{Ours (Single-Trans)} on the left and 
\textbf{Ours (Trans)} on the right. Dots are the facet embeddings outputted by our models and crosses are their nearby entity pair embeddings}
\label{fig:example}
\end{figure*}

\subsection{Embedding Visualization}
\label{sec:visual}
We visualize the embeddings of sentence patterns and a KB relation from the single embedding model and multi-facet embedding model that perform the best in the RE tasks (i.e., \textbf{Ours (Single-Trans)} and \textbf{Ours (Trans)} in Table~\ref{tb:TAC}). We project the facet embeddings to a 2-dimensional space using multidimensional scaling (MDS)~\cite{borg2005modern} and visualize the embeddings of one KB relation and three related sentence patterns in Figure~\ref{fig:example}. The three sentence patterns are selected from our validation set, so the model is not aware of the entity pairs that actually co-occur with the patterns during training. For each facet embedding, we show two among five of its closest entity pairs to visualize the meaning of the embedding space.\footnote{Notice that our training signal is sparse and noisy and the projection does not necessarily preserve the original distances, so the entity pairs with similar relations might be relatively far away from each other.}


In the single embedding model, the embedding of \textit{org:city\_of\_headquarter} is close to the embedding of \textit{(school, location)} while \textit{``\$ARG1 headoffice in \$ARG2''} is close to \textit{(company, location)} and \textit{``\$ARG1 headquarter in \$ARG2''}. 

In the multi-facet embedding model, some embeddings of \textit{org:city\_of\_headquarter} are closer to \textit{(school, location)} and others are closer to \textit{(company, location)}. In addition to these entity pairs, \textit{``\$ARG1 headoffice in \$ARG2''} and \textit{``\$ARG1 headquarter in \$ARG2''} also co-occur with \textit{(people, location)} and \textit{(people/organization, year)}. Using the visualization of multi-facet embedding, we can understand which facets of \textit{org:city\_of\_headquarter} are similar or dissimilar to \textit{``\$ARG1 headoffice in \$ARG2''}, which cannot be done if all facets are averaged into a single embedding as in the traditional models.

The facet embeddings of \textit{``\$ARG1 is now at \$ARG2''} are close to \textit{(people, organization)} where the organization could be school, sports team, and company. Using multiple embeddings could avoid enforcing the closeness of these entity pairs with different relations. The results also indicate that our model can output reasonable cluster centers despite 
learning from the sparse and noisy training data.
Finally, we can see that if a sentence pattern has fewer facets than $K$, our model learns to output some very similar facet embeddings, which makes the performance less sensitive to the setting of $K$.










\subsection{Relation Extraction}
\label{sec:RE_eval}
We follow the same training data and testing protocol in compositional universal schema (CUSchema) \citep{verga2016multilingual}\footnote{\url{https://github.com/patverga/torch-relation-extraction}} to highlight the benefit of predicting multiple facet embeddings, and the relation extraction step in TAC KBP slot-filling tasks is used to compare the different models.

\begin{table*}[t!]
\centering

\scalebox{0.9}{
\begin{tabular}{|c|ccc|ccc|ccc|}
\hline
\multirow{2}{*}{Method} & \multicolumn{3}{|c|}{TAC 2012 (Validation)} & \multicolumn{3}{|c|}{TAC 2013} & \multicolumn{3}{|c|}{TAC 2014} \\ \cline{2-10}
& Prec & Recall & F1 & Prec & Recall & F1 & Prec & Recall & F1 \\ \hline
USchema* & \textbf{34.8} & 23.7 & 28.2 & \textbf{42.6} & 29.4 & 34.8 & \textbf{35.5} & 24.3 & 28.8 \\
CUSchema (LSTM)* &  27.0 & \textbf{32.7} & 29.6 & 39.6 & 32.2 & 35.5 & 32.9 & 27.3 & 29.8\\
Ours (Single-LSTM) & 25.7 & 21.7 & 23.5 & 30.4 & 26.3 & 28.2 & 22.1 & 20.5 & 21.3  \\
Ours (Single-Trans) &  26.1 & 21.6 & 23.7 & 29.5 & 25.2 & 27.2 & 19.0 & 21.2 & 20.0 \\
Ours (LSTM) &  32.0 & 28.9 & 30.3 & 41.3 & \textbf{33.9} & 37.2 & 34.1 & \textbf{29.5} & \textbf{31.6} \\
Ours (Trans) & 33.6 & 27.7 & \textbf{30.4} & 42.5 & 33.2 & \textbf{37.3} & 34.6 & 28.5 & 31.3 \\ \hline
USchema + CUSchema (LSTM)* & 29.3 & 32.8 & 30.9 & \textbf{41.9} & 34.4 & 37.7 & 29.3 & 34.1 &  31.5 \\
USchema + Ours (LSTM) & 29.2 & 33.7 & 31.3 & 38.1 & \textbf{38.9} & 38.5 & 31.5 & \textbf{34.4} & 32.9 \\ 
USchema + Ours (Trans) & \textbf{30.4} & \textbf{33.9} & \textbf{32.1} & 39.0 & 38.8 & \textbf{38.9} & \textbf{32.0} & 34.0 & \textbf{33.0} \\  \hline
\end{tabular}
}
\caption{Distantly supervised relation extraction using different versions of the universal schema. All numbers are \%. CUSchema refers to compositional universal schema. Trans is an abbreviation of Transformer. The best scores of the single models and ensemble models are highlighted. *The performance of TAC 2013 and 2014 are copied from \citet{verga2016multilingual}. }
\label{tb:TAC}
\end{table*}

\textbf{Setup:} 
The training data for our RE models are prepared by distant supervision without requiring any manually labeled data.  The relations in Freebase \citep{bollacker2008freebase} are mapped to TAC relations (e.g., \textit{org:city\_of\_headquarter}) and the NER tagger and entity linker are run in a raw text corpus. Then, the training data is cleaned using the methods in \citet{roth2013survey}.

During testing, we are given a query containing the head entity and a query TAC relation in the slot-filling tasks, and the goal is to extract the tail entity from the candidate sentences. The NER tagger and query expansion are used to gather the candidate sentence patterns, and we compute the similarity scores from different models between the candidate sentence patterns and query relation. Finally, we compare the extracted second entity with the ground truth using exact string matching and report the precision, recall, and F1 scores.

Following \citet{verga2016multilingual}, we use TAC 2012 as our validation set to determine the threshold score for each TAC relation. Each model's hyperparameters are tuned separately using the validation set (TAC 2012) to ensure a fair comparison.

We compare the following methods: \\
\textbf{Ours (Trans)}:
Our method that measures the similarity between the sentence pattern~$\{\tilde{\underline{\vs}}_{i,k}\}$ and TAC relation~$\{\tilde{\underline{\vs}}_{j,m}\}$
using $Sim(\{\tilde{\underline{\vs}}_{i,k}\},\{\tilde{\underline{\vs}}_{j,m}\} )$ in \eqref{eq:sim_score}. Trans is an abbreviation of the Transformer encoder. We set $K=5$ and $K_{rel}=11$ based on the validation set.\\
\textbf{Ours (LSTM)}:
The same as Ours (Trans) except that we use bi-LSTM as our encoder instead. \\
\textbf{Ours (Single-*)}:
Our methods that use single facet embedding to represent each sentence pattern or KB relation. When setting $K=K_{rel}=1$, our decoder becomes the interleaving feedforward layers and cross-attention layers attending to the output embeddings of the encoder.\\
\textbf{CUSchema (LSTM)}: Compositional universal schema \citep{verga2016multilingual}. The method is similar to Ours (Single-LSTM) but uses a different loss function, neural architecture (no decoder), and hyperparameter search procedure. \\
\textbf{USchema}: Universal schema~\citep{riedel2013relation} estimates every sentence pattern embedding by factorizing the co-occurrence matrices (i.e., replacing the bi-LSTM in CUSchema with a look-up table). \\
\textbf{USchema + *}: \citet{verga2016multilingual} show that taking the maximal similarity between USchema and CUSchema model improves the F1. We also apply the same merging procedure to our model.










\textbf{Results:}
In Table~\ref{tb:TAC}, the proposed method \textbf{Ours (Trans)} significantly outperform \textbf{CUSchema (LSTM)} before and after combining with universal schema. As far as we know, our proposed multi-facet embedding is the first method that outperforms compositional universal schema using the same training signal in the distant-supervised RE benchmark they proposed.

Although the recall of \textbf{USchema} is low because it does not exploit the similarity between the patterns (e.g., \textit{``\$ARG1 happily married \$ARG2''} is similar to \textit{``\$ARG1 married \$ARG2''}), \textbf{USchema} has a high precision because it also won’t be misled by the similarity (e.g., \textit{``\$ARG1, and his wife \$ARG2''} expresses the \textit{spouse} relation but \textit{``\$ARG1, his wife, and \$ARG2''} does not)~\citep{verga2016multilingual}. Thus, combining \textbf{USchema} and \textbf{Ours (Trans)} leads to the best performance.

\textbf{Ours (Trans)} and \textbf{Ours (LSTM)} perform similarly. Furthermore, \textbf{Ours (LSTM)} performs much better than \textbf{Ours (Single-LSTM)}, which demonstrates the effectiveness of using multiple embeddings. 
Notice that multiple facet embeddings could improve the performance even after the training data have been cleaned. This indicates that our method is complementary to the noise removal methods in \citet{roth2013survey}.








\begin{table*}[t!]
\centering
\scalebox{0.75}{
\begin{tabular}{|c|c|c|c|c|c|c|}
\hline
Premise (Specific Pattern) & Hypothesis (General Pattern) & Label & Ours & CUSchema & Ours Diff &  Freq Diff  \\ \hline
\$ARG1 , the president of the \$ARG2 & \$ARG1 , the leader of the \$ARG2 & Entailment &  0.98 & 0.94 & + & +  \\
\$ARG1 's chairman , \$ARG2 & \$ARG1 's leader , \$ARG2 & Entailment &  0.95 & 0.87 & + & - \\
\$ARG1 's father , \$ARG2 & \$ARG1 's leader , \$ARG2 & Other &  0.08 & 0.52 & NA & NA  \\
\$ARG1 worked with \$ARG2 & \$ARG1 deal with \$ARG2 & Entailment &  0.92 & 0.83 & + & - \\
\$ARG1 had with \$ARG2 & \$ARG1 deal with \$ARG2 & Other &  0.96 & 0.88 & NA & NA \\
\$ARG1 said the \$ARG2 & \$ARG1 say the \$ARG2 & Paraphrase & 0.93 & 0.92 & NA & NA \\\hline

\end{tabular}
}
\caption{Example of sentence pattern pairs, its label, and our predictions in our entailment experiment. \textbf{Ours} and \textbf{Ours Diff} are the predictions from \textbf{Ours (Trans)}. \textbf{Freq Diff} is the frequency difference baseline. }
\label{tb:entail_examples}
\end{table*}

\subsection{Entailment Detection}
\label{sec:entail_eval}

Entailment is a common and fundamental relation between two sentence patterns. Some examples could be seen in Table~\ref{tb:entail_examples}. Unsupervised hypernym detection (i.e., entailment at the word level) is extensively studied~\cite{shwartz2017hypernyms}, but we are not aware of any previous work on unsupervised entailment detection at the sentence level, nor any existing entailment dataset between sentence patterns. Thus, we create one. 



\textbf{Dataset Creation:}
We use WordNet~\citep{miller1998wordnet} to discover the entailment candidates of sentence pattern pairs and manually label the candidates. For each sentence pattern in the training data of \citet{verga2016multilingual}, we replace one word at a time with its hypernym based on the WordNet hierarchy. The two sentence patterns before and after replacement form an entailment candidate. 

We label 1,500 pairs of the most popular sentence pattern, which co-occurs with the highest number of unique entity pairs. Each candidate could be labeled as entailment, paraphrase, or other. Finally, around 20\% of the candidates are randomly chosen to form the validation set, and the rest are in the test set. More details of the dataset creation process could be seen in \refAppendixEntail


In this dataset, only 22\% and 10\% of candidates are labeled as entailment and paraphrase, respectively. This suggests that entailment relation between two sentence patterns is hard to be inferred by only the hypernym relation (i.e., entailment relation at the word level) in WordNet.







\textbf{Setup:}
We evaluate entailment detection using the typical setup and metrics in hypernym detection \citep{shwartz2017hypernyms}. Negative examples include the candidates labeled as paraphrases and others. We compare the average precision of different methods (i.e., AUC in the precision-recall curve)~\citep{hastie2009elements}. In addition, we predict the direction of entailment relation in each pair (i.e., which pattern is the premise) and report the accuracy. Many hypotheses have the same hypernyms
such as the \textit{leader} in Table~\ref{tb:entail_examples}, so we also report the macro accuracy of direction detection averaged across every hypernym in the hypotheses.






The task is challenging because all the candidates have a word-level entailment relation if their compositional meaning is ignored. Furthermore, we cannot infer the entailment direction based on the tendency that longer sentence patterns tend to be more specific because most of the candidate pairs in this dataset have the same length.


As described in Section~\ref{sec:scoring}, our models detect the direction by computing 
\textbf{Ours Diff} as $Asym(\{\tilde{\underline{\vs}}_{i,k}\},\{\tilde{\underline{\vs}}_{j,m}\} ) - Asym(\{\tilde{\underline{\vs}}_{j,m}\}, \{\tilde{\underline{\vs}}_{i,k}\} )$ and
predict the $i$th sentence pattern to be 
premise if \textbf{Ours Diff} $> 0$. When performing entailment classification, we use as the asymmetric similarity scores $Asym(\{\tilde{\underline{\vs}}_{i,k}\},\{\tilde{\underline{\vs}}_{j,m}\} )$. We report the performance of \textbf{Ours (Trans)}, which is the same best model in the RE experiment.


In entailment classification, we compare the results with cosine similarity from \textbf{Ours (Single-Trans)} and \textbf{CUschema}. 
We also test the frequency difference, which is a strong baseline in hypernym direction detection~\cite{chang2018distributional}. \textbf{Freq Diff} = Freq($S_j$) - Freq($S_i$) where Freq($S_i$) is the number of unique entity pairs co-occurred with the $i$th sentence pattern. The baseline predicts $S_i$ to be premise if \textbf{Freq Diff} $> 0$ because more general sentence patterns should co-occur with more entity pairs. As a reference, we also report the performance of random scores.







\begin{table}[t!]
\centering

\scalebox{0.8}{
\begin{tabular}{|c|c|cc|}
\hline
\multirow{2}{*}{Method} & Classification & \multicolumn{2}{|c|}{Direction Detection}  \\ \cline{2-4}
& AP@all & Micro Acc & Macro Acc \\ \hline
Random & 21.9 & 50.0 & 50.0\\
Freq Diff & 21.4 & 54.5 & 47.3 \\
CUSchema & 31.2 & 50.0 & 50.0\\
Ours (Single) & 23.6 & 50.0 & 50.0 \\
Ours& \textbf{37.8} & \textbf{63.1} & \textbf{55.4} \\ \hline
\end{tabular}
}
\caption{Comparison of entailment detection methods. AP and Acc are average precision and accuracy, respectively. All numbers are \%. Our methods use a Transformer as their encoder. 
}
\label{tb:entailment}
\end{table}

\textbf{Results:}
The quantitative and qualitative comparison are presented in Table~\ref{tb:entailment} and Table~\ref{tb:entail_examples}, respectively. Our model that uses multi-facet embeddings significantly outperforms the other baselines. We hypothesize that a major reason is that the sentence patterns with an entailment relation are often similar in some but not all of the facets, and our asymmetric similarity measurement is better at capturing the facet overlapping.


\section{Related Work}

Relation extraction (RE) is widely studied. \citet{han2020more} summarize the trend of recent studies and point out one of the major challenges is the cost of collecting the labels. Distant supervision~\citep{mintz2009distant} and its follow-up work enable us to collect a large amount of training data at a low cost, but the violation of its assumptions often introduces substantial noise into the supervision signal. Our goal is to alleviate the noise issue by representing every sentence pattern using multiple embeddings. 

Other noise reduction methods have also been proposed~\cite{roth2013survey}. For instance, we can adopt multi-instance learning techniques~\cite{yao2010collective, surdeanu2012multi,amin2020data}, global topic model~\cite{alfonseca2012pattern}, or both~\cite{roth2013feature}. We can also reduce the noise by counting the number of shared entity pairs between a sentence pattern and a KB relation~\cite{takamatsu2012reducing,su2018global}. Nevertheless, the studies focus on mitigating the noise caused by assuming similarity between the sentence patterns and KB relations that co-occur with the same entity pairs, while our method can also reduce the noise from two sentence patterns sharing the same entity pair. Besides, our method is complementary to popular noise reduction methods because our improvement is shown in the training data that have been cleaned~\citep{verga2016multilingual}.









Our method is conceptually related to some studies for lexical semantics. For example, word sense induction or unsupervised hypernymy detection can be addressed by clustering the co-occurring words~\citep{NeelakantanSPM14,athiwaratkun2017multimodal,chang2018distributional}. However, the clustering-based methods do not apply to RE because the co-occurring matrix for RE is much sparser (see Section~\ref{sec:cluster} for more details).


Finally, our work is inspired by 
\citet{NSD}, but they focus on improving the sentence representation rather than RE. We encourage the facet embeddings to become the centers in Kmeans clustering instead of NNSC (non-negative sparse coding) clustering used in \citet{NSD}, due to its simplicity, efficiency, and better RE performance. Moreover, we discover that an additional regularization described in \refAppendixAuto is crucial to overcome the sparsity challenge in RE. 







\section{Conclusion}

In this work, we address the limitation of representing each sentence pattern using only a single embedding, and our approach improves the distantly-supervised RE performances of compositional universal schema. 

Relying on only a very sparse co-occurrence matrix between the sentence patterns and entity pairs, we show that it is possible to predict reasonable cluster centers of entity pair embeddings and to predict the entailment relation between two sentence patterns without any labels.

\section*{Acknowledgements}
We thank Ge Gao for the preliminary exploration of this project. We also thank the anonymous reviewers for their constructive feedback.

This work was supported in part by the Center for Data Science and the Center for Intelligent Information Retrieval, in part by the Chan Zuckerberg Initiative under the project Scientific Knowledge Base Construction, in part using high performance computing equipment obtained under a grant from the Collaborative R\&D Fund managed by the Massachusetts Technology Collaborative, in part by the National Science Foundation (NSF) grant numbers DMR-1534431 and IIS-1514053. 

Any opinions, findings, conclusions, or recommendations expressed in this material are those of the authors and do not necessarily reflect those of the sponsor.




\bibliography{eacl2021}
\bibliographystyle{acl_natbib}

\newpage
\appendix

\section{Method Details}
\label{sec:method_details}


Our objective function includes the loss defined in equation 2 and a regularization term $\Omega$. We describe $\Omega$ in Section~\ref{sec:auto} and some implementation details in Section~\ref{sec:implementation}. Finally, our training procedure is summarized in Algorithm~\ref{algo:dist_opt}.

\subsection{Regularization by Autoencoder}
\label{sec:auto}

The co-occurrence matrix between the sentence patterns and entity pairs is very sparse because most of the sentence patterns only co-occur with a few entity pairs. The sparsity might make the training process of multi-facet embeddings sensitive to the hyperparameters. 


We discover that adding a simple autoencoder regularization is an effective way to stabilize the training. This regularization term aims to make the average of our facet embeddings of a sentence pattern similar to the weighted average of our word embeddings in that sentence pattern. The regularization is a kind of autoencoder because 
it reconstructs the weighted average embeddings of words in the input sentence pattern using the output facet embeddings.
The regularization term $\Omega$ is defined as 

\vspace{-3mm}
\small
\begin{align}
\gamma \sum_{ (i,j) \in R^{auto} } (2 \cdot \mathbbm{1}_{i=j}-1)|| \underline{\vs}^{aw'}_j - \mu(S_i) ||^2,
\label{eq:auto_dist}
\end{align}
\normalsize
where $\gamma$ is a weight for the regularization term, 
$R^{auto} = \cup_{i=1}^I \{(i,i),(i,q)\}$ is the set of all positive and negative training pairs, $I$ is the number of sentence patterns, and $q$ is a randomly selected index of sentence patterns which serves as our negative example, $\mathbbm{1}_{i=j} = 1$ if $i=j$ and $0$ otherwise, 
$\mu(S_i)= \frac{\sum_{k=1}^K \underline{\vs}_{i,k}}{K}$ is the average of facet embeddings of the sentence pattern $S_i$.
$\underline{\vs}^{aw'}_j$ is a weighted average embedding of words in the $j$th sentence pattern that passes through a linear transformation $\bm{H}$.
Weighting each word embedding by a smoothed inverse frequency provides a better text similarity measurement~\citep{arora2016simple} because the frequently occurring words often do not contribute much 
to the semantic meaning (e.g., stop words). Similarly
, we compute
\begin{align}
\underline{\vs}^{aw'}_j = \bm{H} \underline{\vs}^{aw}_j = \bm{H} \sum_{w \in S_j} \frac{\nu}{\nu + p(w)} \underline{\vw}, 
\label{eq:avg_word_emb}
\end{align}
where $\bm{H}$ linearly transforms 
the word embedding into the entity pair embedding space. $\nu=10^{-4}$ is a constant set suggested in \citet{arora2016simple}, $p(w)$ is the frequency of 
the word $w$ divided by the number of words in the corpus, and $\underline{\vw}$ is the pretrained embedding of word $w$. We use 840B GloVe~\cite{glove} as our word embedding in this work.

\begin{table}[t!]
\centering
\scalebox{0.9}{
\begin{tabular}{|c|c|c|}
\hline
Ours (Single-Trans) & Ours (Trans) & BERT Base \\ 
1.0M & 1.1M & 86.0M \\ \hline 
Ours (Single-LSTM) & Ours (LSTM)  \\ 
2.6M & 3.0M \\ \cline{1-2}
\end{tabular}
}
\caption{Comparison of neural model sizes (i.e., the number of trainable parameters excluding the word embedding layer).}
\label{tb:model_size}
\end{table}

\begin{algorithm*}[!thp]
\SetAlgoLined
\SetKwInOut{Input}{Input}
\SetKwInOut{Output}{Output}

\Input{Sentence patterns and KB relations $S$, co-occurrence matrix $\{y_{i,j}\}$, entity pair embeddings from USchema, and pre-trained word embeddings.}
\Output{Our neural encoder and decoder}
Initialize the entity pair embeddings using the embeddings learned by USchema. \\
Initialize the word embeddings of our neural encoder using pre-trained word embeddings and randomly initialize other parameters\\
\ForEach{$S_i$ \text{in training corpus}}{%
	Run forward pass on the neural encoder and decoder to compute facet embeddings $\{\underline{\vs}_{i,k}\}_{k=1}^K$ \\ 
	Collect positive examples (i.e., $\{j|y_{i,j}=1\}$) and negative examples for the $i$th sentence pattern or KB relation\\
	\ForEach{\text{Positive and negative examples} $(i,j)$}{%
	Compute $\tilde{\underline{\ve}}_j= \frac{\underline{\ve}_j}{||\underline{\ve}_j||}$ \\
	Compute $\eta_k=\min(1,\max(0,\frac{\tilde{\underline{\ve}}^T_j \underline{\vs}_{i,k} }{||\underline{\vs}_{i,k}||^2})), \forall 1 \leq k \leq K$ \\
	Select $k_{best}$th facet embedding by $k_{best}=\argmin_{k} || \tilde{\underline{\ve}}_j - \eta_{k} \underline{\vs}_{i,k} ||^2$ \\
	Add $(2 \cdot y_{i,j}-1) r_{i,j} || \tilde{\underline{\ve}}_j - \eta_{k_{best}} \underline{\vs}_{i,k_{best}} ||^2$ to the loss
	}
	Add the autoencoder loss $\Omega$ in \eqref{eq:auto_dist} using pre-trained word embeddings\\
    Update neural encoder and decoder, entity pair embeddings, and $\bm{H}$ by backpropagation
}
 \caption{Training procedure (using batch size = 1)}
 \label{algo:dist_opt}
\end{algorithm*}

\subsection{Implementation Details}
\label{sec:implementation}

In our Transformer encoder and decoder, we set the number of layers to be 3 and the size of the hidden state to be 300. In our bi-LSTM encoder, the number of layers is set to 2. After the bi-LSTM encoder, we use the last embedding in the hidden state to encode a sequence into a single embedding. 
In Table~\ref{tb:model_size}, we can see that the size of the neural model outputting multi-facet embeddings is similar to the size of the single embedding model, and both are much smaller than the BERT base model~\citep{BERT}. The size of CUSchema is smaller than ours because it uses a smaller hidden state size, but we find that increasing the model size of CUSchema does not lead to better performance.





Before training, we initialize the embeddings of entity pairs using USchema as in \citet{verga2016multilingual}. Besides, we initialize the word embedding layer in our neural model using GloVe. CUSchema initializes its word embedding layer randomly, but we also find that initializing it using pre-trained word embeddings does not increase the performance.

We use Adam~\cite{kingma2014adam} to optimize the parameters of our neural model and entity pair embeddings. For the linear layer $\bm{H}$ in \eqref{eq:avg_word_emb}, we adopt SGD to make the training more stable. Due to its small model size, one 1080Ti GPU is sufficient to train our model in 3 days.


\section{Experiment Details}

The details of preparing the entailment dataset are included in Section~\ref{sec:entailment_detail}, and the details of our experiment setup are included in Section~\ref{sec:setup_detail}. We present the results of our ablation studies in Section~\ref{sec:ablation_detail}.

\subsection{Entailment Dataset Creation}
\label{sec:entailment_detail}
When finding the entailment candidates using WordNet, we iterate over all the words in every sentence pattern. For each word, we retrieve all of its senses/synsets and the possible hypernym synsets. Finally, we replace the word with every lemma of each hypernym synset. After the replacement, if the sentence pattern appears in our training data, we pair the sentence patterns before and after replacement as a candidate.




Our goal is to find entailment rather than paraphrase relation, so we exclude the candidates where the replaced word is both the hypernym and hyponym of the replacing word.
To measure each sentence pattern's popularity, we compute the number of unique entity pairs co-occurring with the sentence pattern as the pattern frequency and take the minimum of the frequency between the two sentence patterns in a candidate as the candidate frequency. 
For each hypothesis, we only consider the top 6 premise candidates with the highest frequencies to diversify the hypotheses in our dataset. The hypothesis popularity score is the average candidate frequencies across its top 6 premises.

Before labeling, we sort the hypotheses based on their popularity scores and label the most popular 1,500 candidates (with the highest minimal frequencies). The labeling is done by a PhD student who has RE research experiences because we hypothesize that it is hard to clearly explain the task to crowdsourcing workers. After the dataset is built, we separate the validation set and test set such that all hypernyms in the test hypotheses are unseen in the validation set.








%

\subsection{Experiment Setup Details}
\label{sec:setup_detail}

In our visualization experiment, we filter out a few entity pairs that co-occur with less than 5 sentence patterns or become far away from its closest facet embedding after the projection. To prevent the facet embeddings from overlapping, we add a small random vector to each facet embedding.

In the co-occurrence matrix, we use 5\% of the unique sentence patterns as our validation set. All the sentence patterns in the validation set are unseen in the training set. We use the validation set to tune the number of epochs during training.

We use coordinate descent to search the hyperparameters that result in the best F1 score in TAC 2012 during training (i.e., change one hyperparameter at a time). Compared with the grid search used in \textbf{CUSchema}, this tuning method is less computationally expensive and less likely to overfit the validation data. 

We first optimize the hyperparameters in \textbf{Ours (Trans)}. The search ranges are $\gamma=[0.1, 0.2, 0.3]$, $K=[1, 2, 3, 4, 5, 6, 11]$, $K_{rel}=[1, 8, 9, 10, 11, 12, 13, 14, 15]$, encoder dropout rate $=[0.25, 0.3, 0.35]$, learning rate for updating $\bm{H}=[1, 0.1, 0.01]$ and maximal epoch number $=[15, 20, 25, 30, 50]$. Our best Transformer model \textbf{Ours (Trans)} used $\gamma=0.2$, $K=5$, $K_{rel}=11$, encoder dropout rate $= 0.3$, learning rate for updating $\bm{H}=0.1$ and maximal number of epochs = 50. Then, we start from these best hyperparameters for \textbf{Ours (Trans)} and tune only encoder dropout rate $=[0.25, 0.3, 0.35]$, and maximal epoch number $=[15, 20, 25, 30, 50]$ for \textbf{Ours (LSTM)}. The best performing LSTM model used maximal 30 epochs while all other hyperparameters are found to be the same as the best Transformer model. Finally, we fix $K=K_{rel}=1$ and tune the hyperparameters using the same range as above, for \textbf{Ours (Single-Trans)} and \textbf{Ours (Single-LSTM)}. 





\subsection{Ablation Study}
\label{sec:ablation_detail}

In Table~\ref{tb:TAC_ablation}, we justify using the autoencoder loss and using different facet numbers for sentence patterns ($K$) and for KB relations ($K_{rel}$). We can see that performance drops if we remove these techniques from our models using a Transformer.


\begin{table}[t!]
\centering

\scalebox{0.9}{
\begin{tabular}{|c|c|c|c|}
\hline
Method & 2012 & 2013 & 2014 \\ \hline
Ours & \textbf{30.4} & \textbf{37.3} & \textbf{31.3} \\ 
Ours ($K=K_{rel}=11$) & 29.9 & 36.1 & 29.8 \\
Ours (No autoencoder) & 27.5 & 33.5 & 30.2 \\ \hline
\end{tabular}
}
\caption{Ablation study on TAC datasets. All numbers are F1 (\%). All models use the Transformer encoder. }
\label{tb:TAC_ablation}
\end{table}

\appendix


\end{document}